\documentclass{IOS-Book-Article}

\usepackage{mathptmx}
\usepackage{amsmath}
\usepackage{graphicx}
\usepackage{subcaption}
\usepackage{bm}

\begin{document}
\begin{frontmatter}              

\title{Masked Clinical Modelling:\\A Framework for Synthetic and Augmented Survival Data Generation}
\runningtitle{IOS Press Style Sample}

\author[A]{\fnms{Nicholas I-Hsien Kuo}%
\thanks{Corresponding Author: Nicholas I-Hsien Kuo; E-mail: n.kuo@unsw.edu.au\newline \hspace*{3mm}[Paper archived, not yet formally submitted.]}},
\author[A]{\fnms{Blanca Gallego}}
and
\author[A]{\fnms{Louisa R Jorm}}

\runningauthor{B.P. Manager et al.}
\address[A]{ Centre for Big Data Research in Health, Faculty of Medicine,\\ 
University of New South Wales, Sydney, NSW, Australia}
ORCiD ID: Nicholas I-Hsien Kuo https://orcid.org/0000-0001-8749-7280

\begin{abstract}
Access to real clinical data is often restricted due to privacy obligations, creating significant barriers for healthcare research. Synthetic datasets provide a promising solution, enabling secure data sharing and model development. However, most existing approaches focus on data realism rather than utility -- ensuring that models trained on synthetic data yield clinically meaningful insights comparable to those trained on real data. In this paper, we present Masked Clinical Modelling (MCM), a framework inspired by masked language modelling, designed for both data synthesis and conditional data augmentation. We evaluate this prototype on the WHAS500 dataset using Cox Proportional Hazards models, focusing on the preservation of hazard ratios as key clinical metrics. Our results show that data generated using the MCM framework improves both discrimination and calibration in survival analysis, outperforming existing methods. MCM demonstrates strong potential to support survival data analysis and broader healthcare applications.
\end{abstract}

\end{frontmatter}

\thispagestyle{empty}
\pagestyle{empty}

\section{Introduction}
Access to real clinical data in healthcare is often restricted due to privacy guidelines and regulations, creating challenges for reproducibility and generalisability in clinical research~\cite{kuo2022health, nicholas2024enriching}. Consequently, there is growing interest in realistic synthetic datasets that allow research while protecting patient confidentiality. While synthetic healthcare data generation has advanced, much of the focus has been on visual and statistical realism, often neglecting \textbf{utility} -- the ability of models trained on synthetic data to produce results consistent with those trained on real data. Few studies have assessed the utility of synthetic datasets for survival analysis~\cite{norcliffe2023survivalgan}, a key method used in clinical research, including clinical trials and observational studies. This method often uses Cox Proportional Hazards (CoxPH) models~\cite{cox1972regression} to estimate hazard ratios (HRs), which quantify the impact of variables on time-to-event outcomes like disease remission or death.

To address this gap, we propose \textbf{Masked Clinical Modelling (MCM)}, a novel framework inspired by masked language modeling, as used in models like BERT~\cite{devlin2018bert}. MCM supports both data synthesis and conditional data generation, enabling data augmentation as well. Data synthesis creates realistic datasets for secure sharing, while data augmentation generates targeted data for specific subgroups (e.g., patients of certain sex and age), addressing dataset imbalances. This paper evaluates the MCM prototype using the well-established WHAS500 dataset~\cite{goldberg1988incidence}.

\section{Methods}
\subsection{The WHAS500 Dataset}

The WHAS500 dataset is a subset of the broader Worcester Heart Attack Study (WHAS)~\cite{goldberg1988incidence}, a long-term, population-based study aimed at understanding the incidence and survival outcomes following acute myocardial infarction. WHAS500 consists of 500 patients and is widely used for time-to-event analysis. To evaluate the performance of our MCM prototype, we focused on the following key features:

\begin{itemize} 
    \item \textbf{Demographics}: Age and sex. 
    \item \textbf{Measurements}: Body mass index (BMI) and systolic blood pressure (SBP). 
    \item \textbf{Chronic Conditions}: History of atrial fibrillation (AF) and\\\hspace*{45.75mm} congestive heart failure (CHF). \end{itemize}

The baseline is a male patient with average BMI, normal SBP, and no history of AF or CHF. The primary endpoint is death. See detailed variable descriptions in \cite{hosmer2008applied}.

\subsection{Masked Clinical Modelling}\label{Sec:MCM}

\begin{figure}[h]
    \centering
    \includegraphics[width=\linewidth]{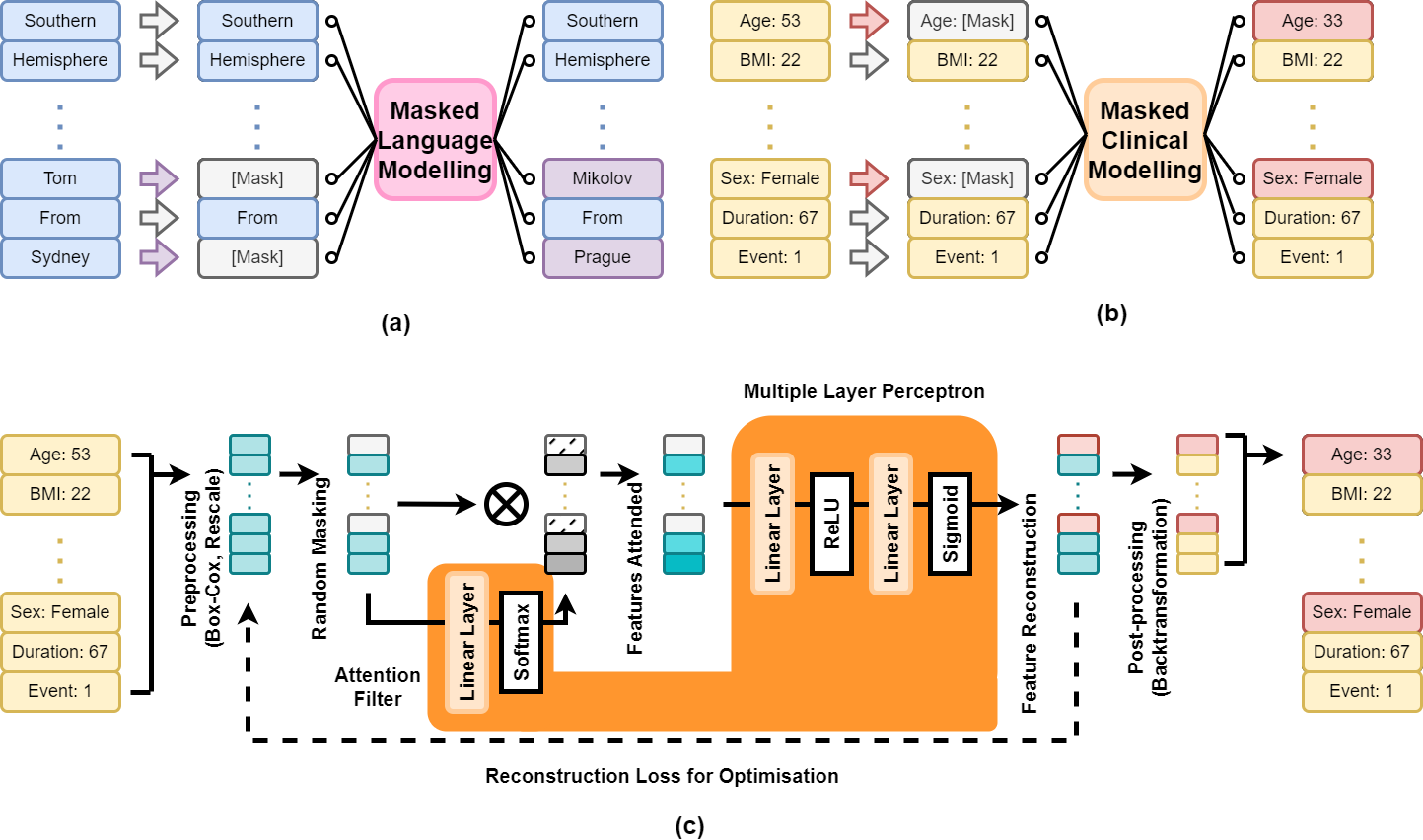}
    \caption{An overview of our masked clinical modelling framework.}
    \label{fig:Pipeline}
\end{figure}

Figure \ref{fig:Pipeline} provides an overview of our Masked Clinical Modelling (MCM) framework. In subfigure (a), masked language modelling hides random words in a sentence, predicting the missing words using context. Subfigure (b) adapts this to \textbf{MCM}, where a patient's clinical data $\bm{x}$ with $N$ features is randomly masked (\textit{e.g.,} age, sex), and the model reconstructs the missing information from the remaining data. Subfigure (c) shows the engineering pipeline: pre-processing transforms $\bm{x}\rightarrow \bm{v}$ via Box-Cox transformation~\cite{box1964analysis} to normalise numeric variables and rescale all features to $[0, 1]$. After masking, the data is passed to the model.

The MCM backbone includes an \textbf{Attention Filter (AF)} and a \textbf{Multi-Layer Perceptron (MLP)}, represented as $F_\text{MCM} = F_\text{AF}\circ F_\text{MLP}$. The AF applies a linear layer with weights $\bm{W}$ followed by a softmax, generating an attention vector $\bm{a}$, where the entries sum to 1. The attention values represent the importance of observed features for reconstructing the masked ones, with masked features receiving an attention score of 0:\\
\hspace*{5mm}$\bm{a} = F_\text{AF}(\bm{v}) = \text{softmax}(\bm{Wv}), \quad \bm{a}_i \in [0, 1], \quad \sum_{i=1}^{N} \bm{a}_i = 1, \quad \bm{a}_i = 0 \ \text{if} \ \bm{x}_i \ \text{is masked}$.\\
The weighted data $\bm{a} \odot \bm{v}$ is passed a two-layer MLP with weights $\bm{U}$:\\
\hspace*{5mm}$\hat{\bm{v}} = F_\text{MLP}(\bm{a} \odot \bm{v}) = \text{sigmoid}\left(\bm{U}_2\left(\text{ReLU}\left(\bm{U}_1(\bm{a} \odot \bm{v})\right)\right)\right)$.\\
The first layer uses a ReLU activation to capture complex patterns, and the second applies a sigmoid activation to ensure the output $\hat{\bm{v}}$ remains in $[0, 1]$.

The model is trained to minimise the mean squared error:
$L = \frac{1}{N}\sum_{i=1}^{N}(\hat{\bm{v}}_i - \bm{v}_i)^2$.
Once trained, the MCM model imputes missing data. Post-processing reverses the Box-Cox transformation and rescaling, restoring the features to their original distributions ($\hat{\bm{v}}\rightarrow\hat{\bm{x}}$).

\newpage
This approach supports both data synthesis and augmentation. For synthesis, parts of an existing dataset are masked and reconstructed. For augmentation, patient profiles with specific traits can be generated. For example, to generate 300 patients aged 50--55 with follow-up times of 240--300 days, we create a data frame with 300 rows, sample the desired values from uniform distributions, and pass the incomplete data to MCM for completion. This enables the generation of clinically relevant data for downstream tasks.

We present a prototype MCM model with a two-layer MLP and 64 hidden dimensions. Due to space constraints, selected results are presented in the Results section, with a more detailed analysis planned for future work. Upon acceptance, all code will be made publicly available via GitHub.

\section{Results}
\begin{figure}[t]
    \centering
    \begin{subfigure}[b]{0.475\textwidth}
        \centering
        \includegraphics[width=\textwidth]{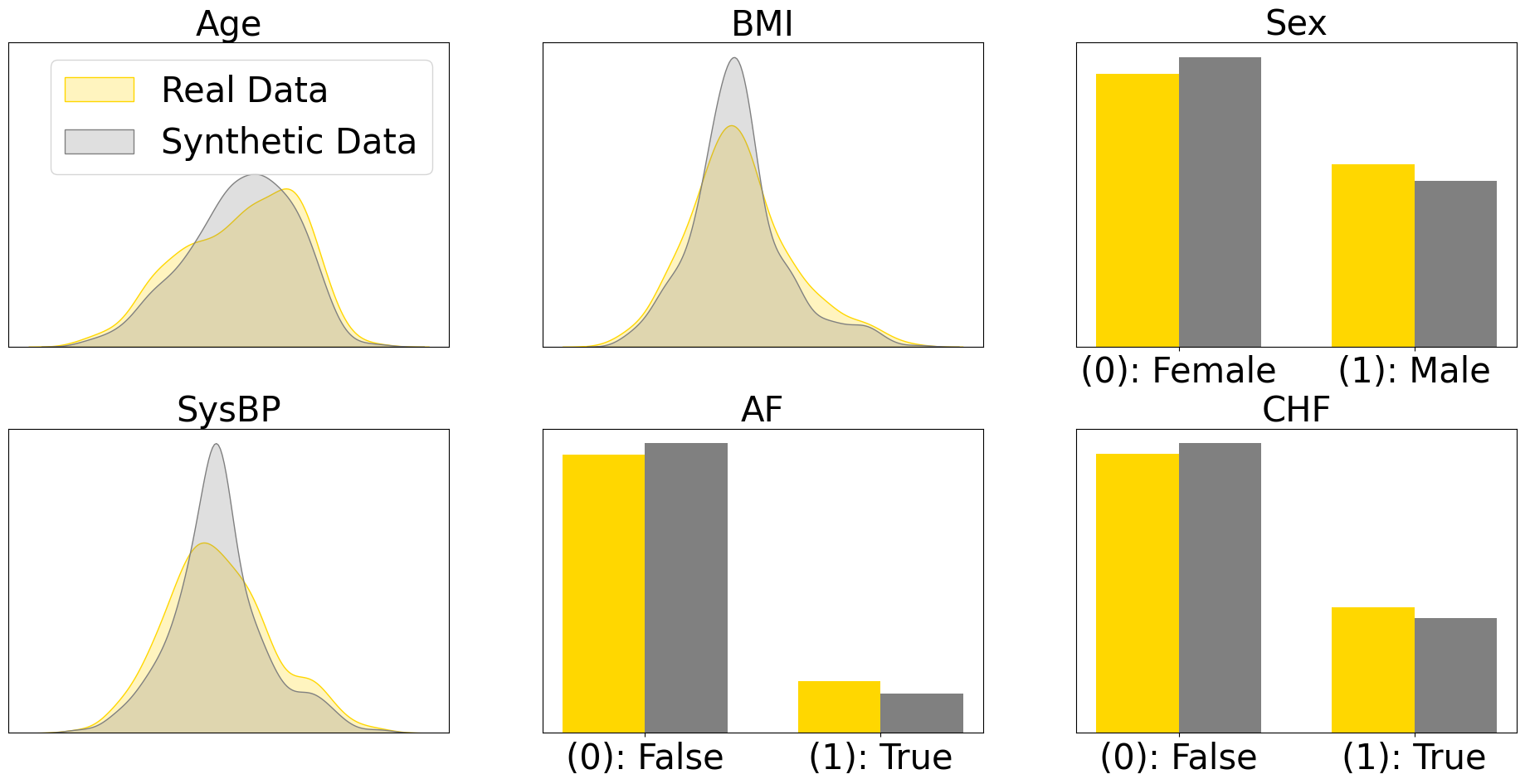}  
        \caption{Distribution Comparison}
        \label{fig:subfig1}
    \end{subfigure}
    \hfill  
    \begin{subfigure}[b]{0.475\textwidth}
        \centering
        \includegraphics[width=\textwidth]{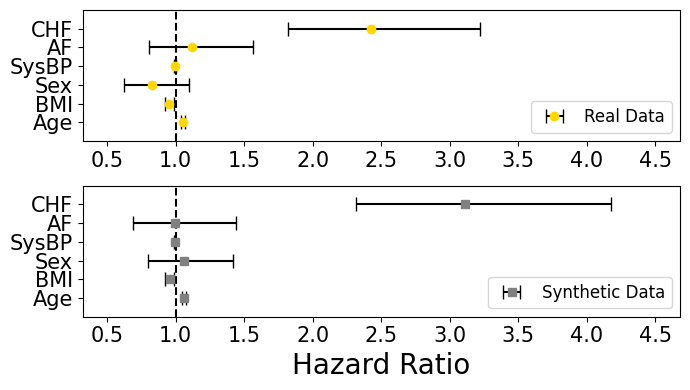}  
        \caption{HR Consistency Comparison}
        \label{fig:subfig2}
    \end{subfigure}
    \caption{Comparing the synthetic data from MCM (in yellow) to the ground truth (in black).}
    \label{fig:SyntheticDataComp}
\end{figure}

In Figure \ref{fig:SyntheticDataComp}, we show how MCM generates synthetic data that closely mirrors the ground truth. After masking 75\% of the real data, the model reconstructed the missing values. Subfigure (a) compares numeric distributions and binary variables, showing a strong match between real and synthetic data. To evaluate clinical utility, we trained a CoxPH model on both datasets, and compared the hazard ratios (HRs). Subfigure (b) demonstrates strong HR alignment, indicating that the synthetic data conserves meaningful correlations for downstream analysis.

We explore how synthetic data can augment ground truth to enhance the performance of a downstream CoxPH model, focusing on two key metrics: discrimination, measured by Harrell’s C-index~\cite{harrell1982evaluating} (higher is better, perfect discrimination is 1), and calibration, assessed by deviation from a perfect slope of 1 (lower is better). Results, summarised in Table \ref{tab:SyntheticComparison}, are based on a 5x2 cross-validation~\cite{dietterich1998approximate}, with mean (SD) reported. Baseline comparisons include CoxPH models trained solely on real data and those augmented with synthetic data from SMOTE~\cite{chawla2002smote}, VAE~\cite{kingma2014auto}, MICE~\cite{van2011mice}, and our MCM method. SMOTE detects minority classes and generates data through linear interpolation; VAE reconstructs data using a neural network with sampled latent features; and MICE imputes missing data using Bayesian ridge regression after masking~\cite{mackay1992bayesian}. All synthetic data are added to the training data for each cross-validation fold.

\begin{table}[h]
    \centering
    \caption{Discrimination, and stratified calibration for two specific cohorts}
    \begin{tabular}{lccccc}
        \hline
        \textbf{} & \textbf{Discrimination} &
        \multicolumn{2}{c}{\textbf{Calibration}} & \multicolumn{2}{c}{\textbf{Calibration}}\\
        \textbf{} & \textbf{} &
        \multicolumn{2}{c}{\textbf{Age $>$ 75}} & \multicolumn{2}{c}{\textbf{Hypertension Stage 2}}\\
        \hline 
        \textbf{Percentile}
        & \textbf{- -}
        & \textbf{25th} & \textbf{75th}
        & \textbf{25th} & \textbf{75th} \\ 
        \hline
        \textbf{Real Data}
        & 0.7609 (0.0222)
        & 0.97 (0.83) & 0.24 (0.25) 
        & 0.86 (0.30) & 0.24 (0.17) \\ 
        \hline
        \textbf{SMOTE}
        & 0.7640 (0.0176)
        & 0.49 (0.38) 
        & 0.27 (0.24) 
        & \textbf{0.19} (0.13) 
        & \textbf{0.10} (0.03) \\
        \textbf{VAE}
        & 0.7609 (0.0185)
        & 1.16 (0.81) 
        & 0.39 (0.68) 
        & 0.54 (0.17) 
        & 0.11 (0.08) \\
        \textbf{MICE}
        & 0.7472 (0.0219)
        & 0.86 (0.30) 
        & 0.24 (0.17) 
        & 0.54 (0.27) 
        & 0.18 (0.07) \\ 
        \hline
        \textbf{MCM (Ours)}
        & \textbf{0.7662} (0.0196)
        & \textbf{0.36} (0.18) 
        & \textbf{0.20} (0.16) 
        & 0.33 (0.21) 
        & 0.19 (0.07) \\ \hline
    \end{tabular}
    \label{tab:SyntheticComparison}
\end{table}
For discrimination, VAE, MICE, and MCM each generated 500 novel synthetic patients. MICE and MCM reconstructed data by randomly masking 50\% of features; and SMOTE synthesised 70 additional patients from the minority class. Our MCM method produced the greatest improvement in the C-index, raising the score from 0.7609 to 0.7662. SMOTE provided slight improvement, VAE showed no effect, and MICE degraded the model’s performance.

For calibration, we stratified patients into two cohorts: 216 individuals aged over 75 and 276 with hypertension stage 2 (SBP $>$ 140 mmHg). Calibration was computed at the 25th and 75th percentiles of time-to-event durations to evaluate synthetic data’s impact on risk prediction for high- and low-risk patients. VAE, MICE, and MCM conditionally generated 5 times the data for the stratified cohort; and SMOTE was employed to rebalance the underlying distributions. With real data alone, CoxPH showed poor calibration at the 25th percentile, with slope misalignments of 0.97 for the over-75 cohort and 0.86 for hypertension. MCM significantly reduced misalignments to 0.36 and 0.33, respectively. At the 75th percentile, where CoxPH performed well, MCM further improved calibration. MICE performed reasonably across scenarios, VAEs were good for hypertension but underperformed for the over-75 cohort, and SMOTE yielded mixed results.

\section{Discussion}
MCM introduces a novel approach to synthetic and augmented survival data generation. By adapting masked language modeling, it ensures both realism and utility by preserving key clinical metrics like HRs. Applied to the WHAS500 dataset, MCM demonstrated improved discrimination and calibration for survival analysis, outperforming SMOTE, VAE, and MICE in CoxPH tasks. Its conditional generation capability presents potential to minimise risks of identity disclosure and attribution (\textit{i.e.,} discovering something new about an individual)~\cite{el2020evaluating} in synthetic data by enriching datasets with more individuals having uncommon combinations of characteristics.

\section{Conclusion}
MCM balances data realism and utility, making it valuable for generating synthetic datasets that retain clinical insights. By preserving HR consistency, MCM supports data sharing in clinical research and education.

\end{document}